\documentclass[letterpaper, 10 pt, conference]{ieeeconf}

\usepackage{graphicx}
\usepackage{xcolor}
\definecolor{mydarkblue}{rgb}{0,0.08,0.45}
\usepackage[colorlinks = true,
            linkcolor = mydarkblue,
            urlcolor  = mydarkblue,
            citecolor = mydarkblue,
            anchorcolor = mydarkblue]{hyperref}
\usepackage{amsmath}
\usepackage{amssymb}
\usepackage{mathtools}
\usepackage{bm}
\usepackage{glossaries}

\newcommand{\glsshortonly}[1]{\glslink{#1}{\glsentryshort{#1}}}

\IEEEoverridecommandlockouts
\newacronym{drl}{DRL}{Deep Reinforcement Learning}
\newacronym{er}{ER}{Embodiment Randomization}
\newacronym{mjx}{MJX}{MuJoCo XLA}
\newacronym{ppo}{PPO}{Proximal Policy Optimization}
\newacronym{dr}{DR}{Domain Randomization}
\newacronym{mtrl}{MTRL}{Multi-Task Reinforcement Learning}
\newacronym{gnn}{GNN}{Graph Neural Network}
\newacronym{urma}{URMA}{Unified Robot Morphology Architecture}
\newacronym{urmav2}{URMAv2}{Unified Robot Morphology Architecture v2}
\newacronym{mlp}{MLP}{Multilayer Perceptron}

\title{\LARGE \bf
Multi-Embodiment Locomotion at Scale \\with extreme Embodiment Randomization
}

\author{Anonymous}

\author{
  Nico Bohlinger$^{1}$,
  Jan Peters$^{1,2}$
\thanks{
$^{1}$Technical University of Darmstadt.
$^{2}$hessian.AI. \& German Research Center for AI (DFKI). \& Robotics Institute Germany (RIG).
}}

\begin{document}

\maketitle
\thispagestyle{empty}
\pagestyle{empty}

\begin{abstract}

We present a single, general locomotion policy trained on a diverse collection of 50 legged robots.
By combining an improved embodiment-aware architecture (URMAv2) with a performance-based curriculum for extreme Embodiment Randomization, our policy learns to control millions of morphological variations.
Our policy achieves zero-shot transfer to unseen real-world humanoid and quadruped robots.

\end{abstract}

\begin{figure}[!htbp]
\centering
\includegraphics[width=\linewidth]{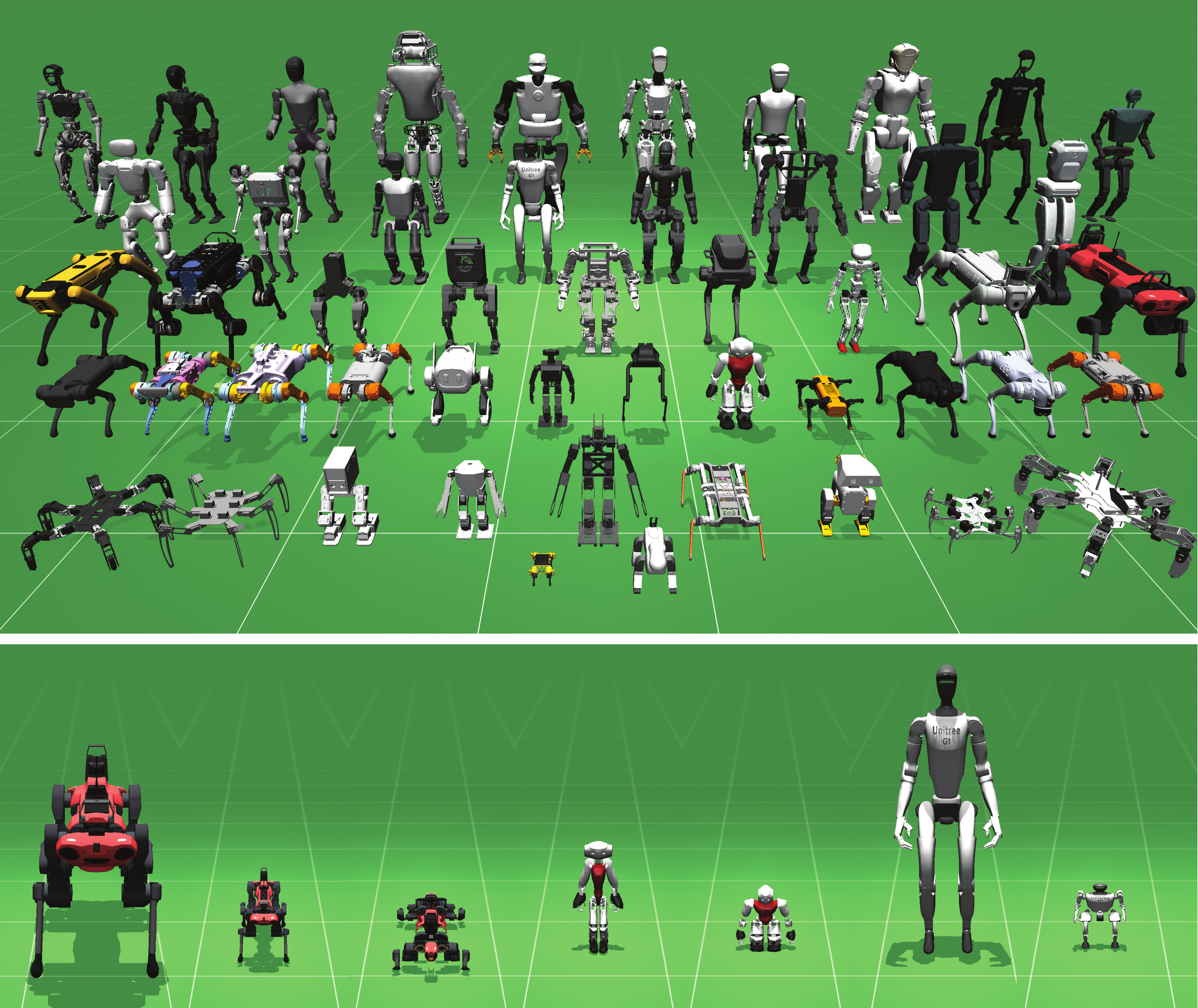}
\caption{
(Top) We collected a diverse set of 50 legged robots, including 15 quadrupeds, 23 humanoids, 8 bipeds and 4 hexapods.
(Bottom) We train the policy on all robots simultaneously using 25600 parallel environments.
The performance-based curriculum on extreme Embodiment Randomization leads to the policy seeing gradually more challenging embodiments throughout training.
This results in a curriculum of up to 10 million different embodiments per training run.
Here different generated varations of the ANYmal C, Nao v5 and Unitree G1 are shown as examples.
}
\label{fig:robots}
\vspace{-0.1em}
\end{figure}

\section{INTRODUCTION}

Recent advances and availability of powerful robot hardware, like humanoid robots, have enabled researchers all around the world to tackle more complex tasks in robotics \cite{ma2025learning, su2025hitter}.
\gls{drl} has shown impressive results in many of these tasks, especially in the field of locomotion \cite{margolis2023, cheng2023}.
With more and more robot platforms being developed and finding their way into research labs and real-world applications, the current paradigm of training a control policy tailored to a specific robot can become increasingly inefficient.
Robot platforms change, adapt, and evolve over time, but many current training approaches do not consider robot morphologies as a key factor.
Their learning process is agnostic or simply unaware of the specific characteristics and capabilities of the robot's embodiment, making cross-embodiment transfer difficult or even impossible.
We build upon the recently introduced \gls{urma} \cite{bohlinger2024onepolicy}, an embodiment-aware learning framework that addresses these challenges for the field of robot locomotion.
We train a single unified embodiment-aware policy across 50 different legged robots with massive \gls{er}.
This results in a curriculum of up to 10 million embodiments per training run, in order to learn a robust and adaptive general locomotion policy, that can be directly zero-shot transferred to unseen humanoid and quadruped robots in the real world.

\section{RELATED WORK}

\begin{figure*}[!htbp]
\centering
\includegraphics[width=0.9\textwidth]{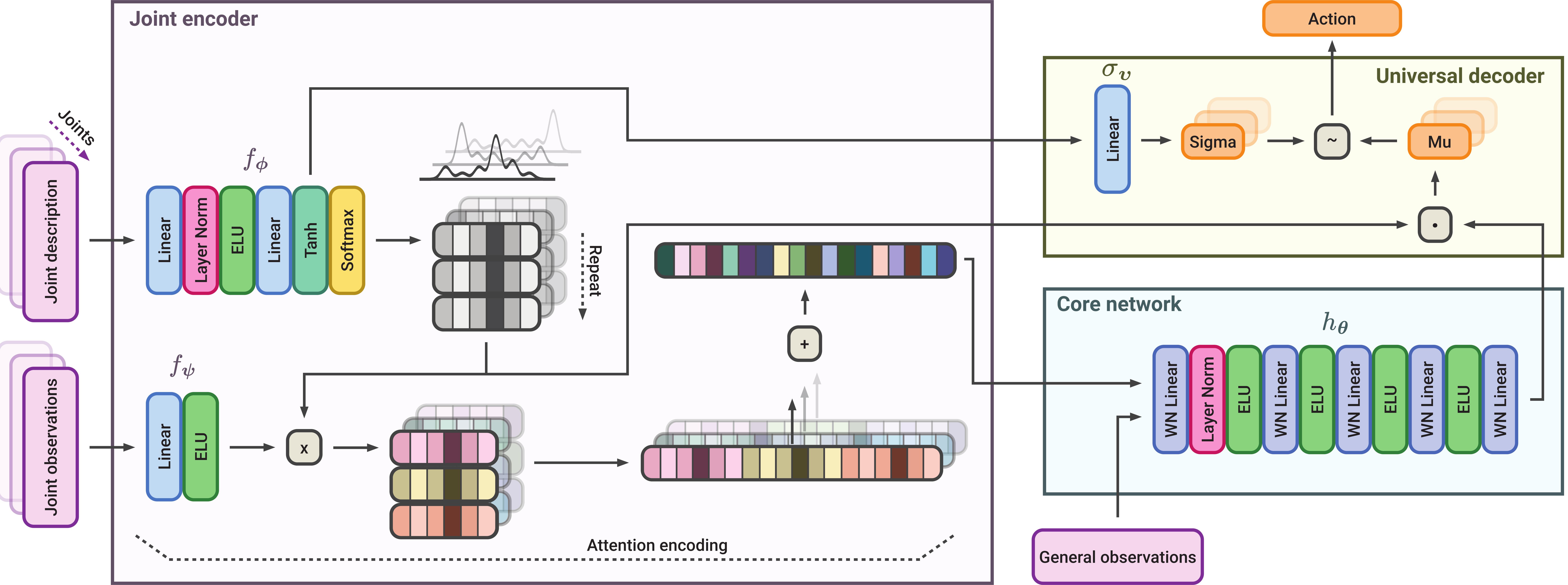}
\caption{
Overview of \glsshortonly{urmav2}.
We extend the original \gls{urma} architecture to improve its scalability, learning stability and empirical performance in the massively multi-embodiment setting.
We increase the capacity of the encoder and core network, add WeightNorm layers for more stable training, and replace the original universal decoder with a streamlined attention-based decoding mechanism.
}
\label{fig:architecture}
\vspace{-0.5em}
\end{figure*}

Robot locomotion has seen significant advancements in recent years, particularly with the rise of \gls{drl} techniques \cite{margolis2023, cheng2023, stasica2025}.
Leveraging fast and highly parallelizable simulators, like Issac Gym/Sim \cite{mittal2023} or \gls{mjx} \cite{todorov2012}, in combination with strong \gls{dr}, and the scalability of on-policy \gls{drl} algorithms, like \gls{ppo} \cite{schulman2017ppo}, has enabled learning robust and high-performing locomotion policies for various quadruped and humanoid robots.
Techniques such as \gls{dr} \cite{tobin2017domain} and student-teacher learning \cite{cheng2023} are used to bridge the sim-to-real gap and other methods, like curriculum learning, help to speed up and stabilize the training process \cite{margolis2024rapid}.
While training directly on the real robot system would be ideal to make the policy fully aware of the true capabilities of its embodiment, it is often impractical due to safety concerns, wear and tear, and the extensive time required for the unparallelized training \cite{bohlinger2025gait}.

In the pursuit of obtaining a foundation model for robotics tasks, like locomotion, that can generalize across different embodiments, the concept of embodiment-aware learning and the technical challenge of different amounts of sensors and actuators (meaning different observation and action spaces in the language of \gls{drl}) come into play.
It is natural to consider the field of \gls{mtrl} here, where a single policy is trained to solve multiple different tasks.
However, many existing \gls{mtrl} approaches simply zero-pad observations and actions or learn different input and output heads for each task, which can severely limit their ability to generalize across different embodiments, as they neglect their structural similarities \cite{bohlinger2024onepolicy}.
Therefore, prior work has proposed \glspl{gnn} as an architecture to better capture the structure of robot embodiments \cite{wang2018}.
Following work has used Transformers in combination with different graph-based features or attention mechanisms to improve the scalability in the multi-embodiment setting \cite{gupta2022}.
Only recently, the \gls{urma} \cite{bohlinger2024onepolicy} could show the applicability of multi-embodiment learning to real-world robots.
While the initial study only used 16 robots and could not show sim-to-real transfer to seen or unseen humanoids, following work has shown embodiment scaling laws for training \gls{urma} on up to 1000 offline generated robots (based on three template morphologies) and demonstrated its transfer to a real humanoid \cite{ai2025towards}.

\section{METHOD}

We build upon the original \gls{urma} architecture and training framework, and scale it to the massively multi-embodiment setting of 50 different base robots with 10 million variations.
To achieve this, we modify the neural network architecture to be larger, more stable and leverage the attention mechanism also for the policy output (see \autoref{fig:architecture}), which we call \gls{urmav2}.
Furthermore, we introduce a performance-based curriculum learning strategy in combination with extreme \gls{er} to expose the policy to gradually more diverse and difficult embodiments.

\subsection{URMAv2 Architecture}

\textbf{Inputs:}
Following the original \gls{urma} architecture, the inputs are split into three categories: per-joint description vectors $\{d_j\}_{j\in\mathcal{J}}$ for the set $\mathcal{J}$ of all joints in a given robot that uniquely describe a joint's static properties (e.g., rotation axis, torque limits), per-joint observations $\{o_j\}_{j\in\mathcal{J}}$ containing dynamic state information (e.g., position, velocity), and general robot observations $\mathbf{o}_g$ (e.g., trunk velocity, gravity vector).
\gls{urmav2} includes an additional per-joint observation to indicate whether a joint should track its nominal position or can be controlled freely, allowing for task-specific joint-level conditioning.
Also, we remove the feet-specific observations and their encoding from the policy network, as contact sensors are not available on many of the considered robots.
The critic network keeps the feet observations and now also receives the noise-free observations to better estimate values.

\textbf{Joint Encoder:}
\gls{urmav2} keeps the same attention-based joint encoder, which processes each joint's description (attention keys) and observation vector (attention values), and aggregates them into the combined joint latent vector
\begin{equation}
\begin{aligned}
    \bar{z}_\text{joints} &= \sum_{j \in \mathcal{J}} z_j,
    \quad 
    z_j = \alpha_j f_\psi(o_j), \\
    \alpha_j &= \frac{\exp\!\left(f_\phi(d_j) / \tau \right)}
                    {\sum_{L_d} \exp\!\left(f_\phi(d_j) / \tau \right)} .
\end{aligned}
\end{equation}

where $f_\phi$ (with latent dimension $L_d$) and $f_\psi$ are the encoders for the joint descriptions and joint observations, respectively, and $\tau$ is the learnable temperature parameter of the softmax.
\gls{urmav2} uses a wider \gls{mlp} for $f_\psi$ (2x 256 units) for the policy network to increase its capacity for the larger number of robots and variations.

\textbf{Core Network:}
The joint latent vector contains all joint information in a fixed-size vector, so it can be concatenated with the fixed-size general observations and processed by the core network $h_\theta$ to generate the action latent vector
\begin{equation}
    \bar{z}_\text{action} = h_\theta(o_g, \bar{z}_\text{joints}).
\end{equation}
\gls{urmav2} uses a deeper stack of 5 instead of 3 hidden layers to increase the model capacity.
To stabilize training, WeightNorm layers \cite{salimans2016weight} are used around every Dense layer, which decompose the weights as $w = g \frac{v}{\|v\|_2}$, where the optimizer acts on $g$, a learnable scalar, and $v$, representing the raw weights.

\textbf{Action Decoder:}
The most significant architectural change is the replacement of \gls{urma}'s universal decoder with an attention-based decoding mechanism.
Instead of concatenating the action latent vector in batch with encoded joint description latents to produce actions for every joint, \gls{urmav2} computes the mean action $\mu_j$ for each joint via a simple dot product between the action latent vector $\overline{\mathbf{z}}_{\text{action}}$ and the corresponding joint's attention weights $\alpha_j$ that were generated in the encoder:
\begin{equation}
\mu_j = \overline{\mathbf{z}}_{\text{action}} \cdot \alpha_j
\end{equation}
Also, the per-joint standard deviations are predicted with a linear layer $\sigma_{\upsilon}$ from the same joint description encoding calculated for the attention weights, which results in actions being sampled from
\begin{equation}
    a_j \sim \mathcal{N}(\mu_j, \sigma_{\upsilon}(f_\phi(d_j))).
\end{equation}
This creates a streamlined architecture that is both conceptually simpler, computationally more efficient and empirically more performant than the original \gls{urma} decoder.

\subsection{Embodiment Randomization}

To improve the generalization capabilities of the policy across different embodiments, we apply extreme \gls{er} online during training (see \autoref{fig:robots}).
\gls{er} differs from standard \gls{dr} in that all the generated values are seen by the policy through the description vectors, allowing the policy to condition and adapt to them.
We use \gls{dr} after the \gls{er} sampling to further modify the sampled parameters but keep them hidden from the policy, to add robustness and improve sim-to-real transfer.
Our \gls{er} includes scaling of: body part size and position in every dimension, coupled and decoupled mass and inertia, center of mass, inertia and body part and joint axis orientation, IMU position, motor torque and velocity and position limits, joint damping and friction and armature and stiffness, joint nominal position, PD gains, and action scaling factor.
Our framework samples a new embodiment during every episode step with a probability of 0.2\% which corresponds to once every 10 seconds of simulated time on average at the highest curriculum level.
This leads to up to 10 million different embodiments per training run.

\subsection{Performance-based Curriculum}

When drastically increasing the number of robots and their variations, the learning problem becomes significantly more challenging.
To tackle this, we introduce a performance-based curriculum learning strategy that attaches every component of the learning framework to a single curriculum coefficient $\beta \in [0, 1]$.
This coefficient is initialized to $\beta=0$ and is increased by $n \Delta \beta$ whenever an episode is deemed successful, e.g., a minimum tracking error, episode length, or return threshold is reached.
$n$ is the number of consecutive successful or unsuccessful episodes, depending on whether $\beta$ should be increased or decreased, and allows the curriculum to quickly adapt to the current performance of the policy.
$\Delta \beta$ is a small constant step size that determines the granularity of the curriculum.
We attach all training components: domain and embodiment randomization ranges, perturbations, sampling probabilities, terrain attributes, termination conditions, and reward penalty coefficients, to this single curriculum coefficient $\beta$.
This significantly helps to speed up the training process for challenging embodiments, especially humanoids, and leads to more stable training runs.

Furthermore, we define all mentioned components in a percentage-based manner based on the robot's nominal parameters from the URDF, which allows us to use the exact same parameters, ranges and reward coefficients for all robots.
For any given robot, only the nominal joint positions and the PD gains have to be specified, the rest of the training framework is fully shared between all robots.

\section{EXPERIMENTS}

\begin{figure}[t]
\centering
\includegraphics[width=\linewidth]{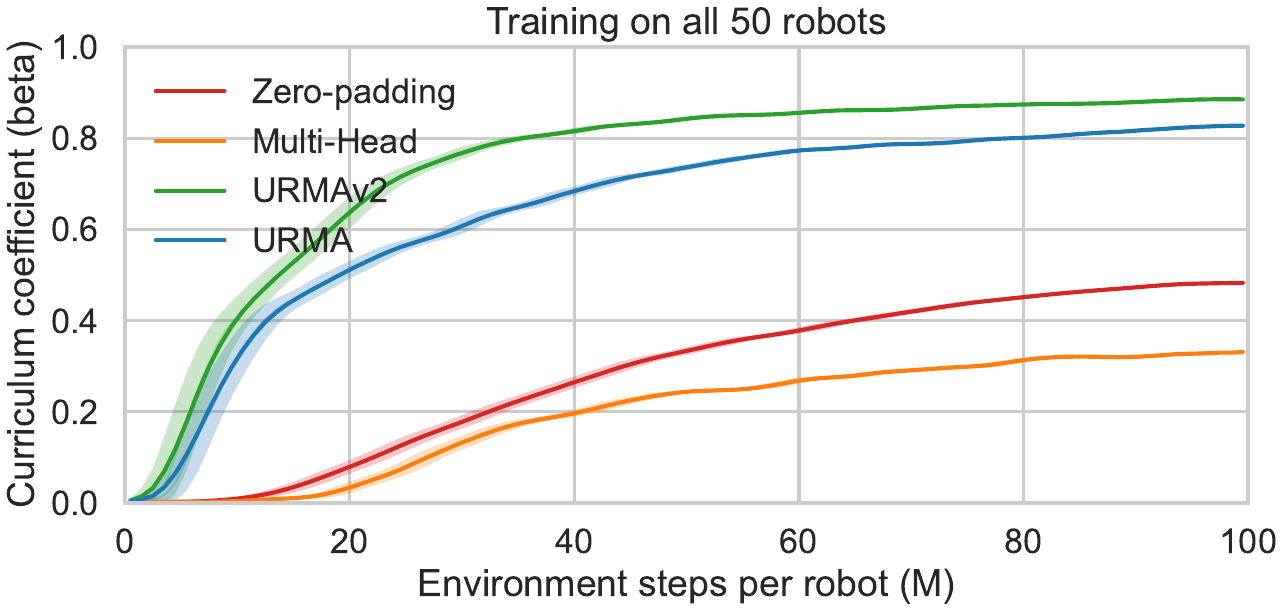}
\caption{
Comparison of the training performance of \gls{urma}, \gls{urmav2}, zero-padding and multi-head baselines when training on all 50 robots.
}
\label{fig:all_50_robots}
\end{figure}

\begin{figure}[t]
\centering
\includegraphics[width=\linewidth]{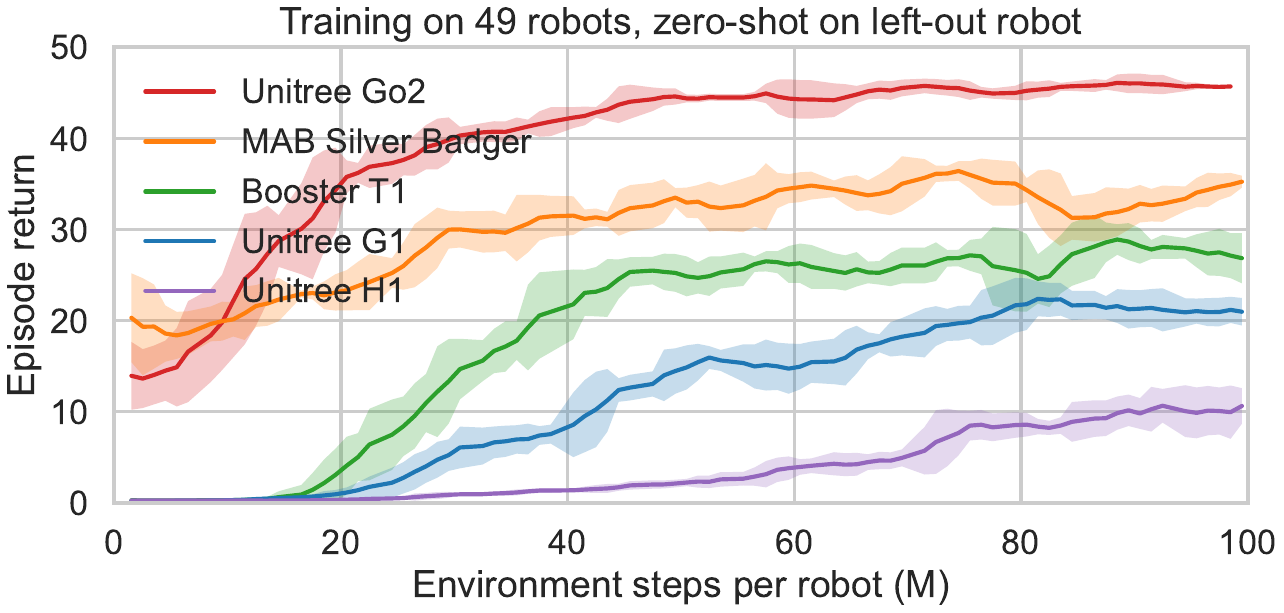}
\caption{
Different setups of \gls{urmav2} trained on 49 robots and zero-shot performance evaluated on the left-out robot.
We test on the MAB Silver Badger and Unitree Go2 for quadrupeds, and the Unitree H1, Unitree G1 and Booster T1 for humanoids.
}
\label{fig:zero_shot_robots}
\end{figure}

We train \gls{urmav2} on a set of 50 legged robots, including 15 quadrupeds, 23 humanoids, 8 bipeds and 4 hexapods, collected from various freely available URDFs (see \autoref{fig:robots}).
We use \gls{mjx} as the physics engine and \gls{ppo} as the \gls{drl} algorithm, which we implement with the RL-X library \cite{bohlinger2023rlx}.
With a total of 25600 parallel environments (512 per robot), we collect 1.6 million samples (32768 per robot) and use 16 minibatches of size 102400 (2048 per robot) for 10 epochs for every policy update.
We train for a total of 5 billion environment steps (100 million per robot), which takes approximately 40 hours on a single NVIDIA A100 GPU.

\autoref{fig:all_50_robots} compares the training performance of \gls{urmav2} with the original \gls{urma} architecture, as well as a zero-padding and multi-head baseline (one head per robot).
We measure the performance by the average curriculum coefficient $\beta$ over all robots.
\gls{urma} and \gls{urmav2} significantly outperform the zero-padding and multi-head baselines, showing the effectiveness of the embodiment-aware architecture in general.
\gls{urmav2} outperforms \gls{urma} in terms of learning speed and final performance ($\beta=0.88$ vs. $\beta=0.82$).

\autoref{fig:zero_shot_robots} shows the zero-shot transfer performance of \gls{urmav2} trained on 49 robots and evaluated on the left-out robot.
For quadrupeds, we test on the Unitree Go2 and MAB Silver Badger (has an additional spine joint), and for humanoids, we test on the Unitree H1, Unitree G1 and Booster T1.
\gls{urmav2} shows strong zero-shot performance on especially the quadruped robots, even the MAB Silver Badger, which proved to be challenging in the original \gls{urma} study due to its additional spine joint that is not present in any other training robot.
Zero-shot performance for the humanoids is clearly lower, but still a significant improvement over the reported 0 return for the Unitree H1 in the original \gls{urma} study.
In simulation, the policy is able to control all three humanoids fairly well, especially the Booster T1 and the Unitree G1, but under server perturbations or really rough terrain, it still falls occasionally.

\subsection{Sim-to-Real Transfer}

\begin{figure}[t]
\centering
\includegraphics[width=\linewidth]{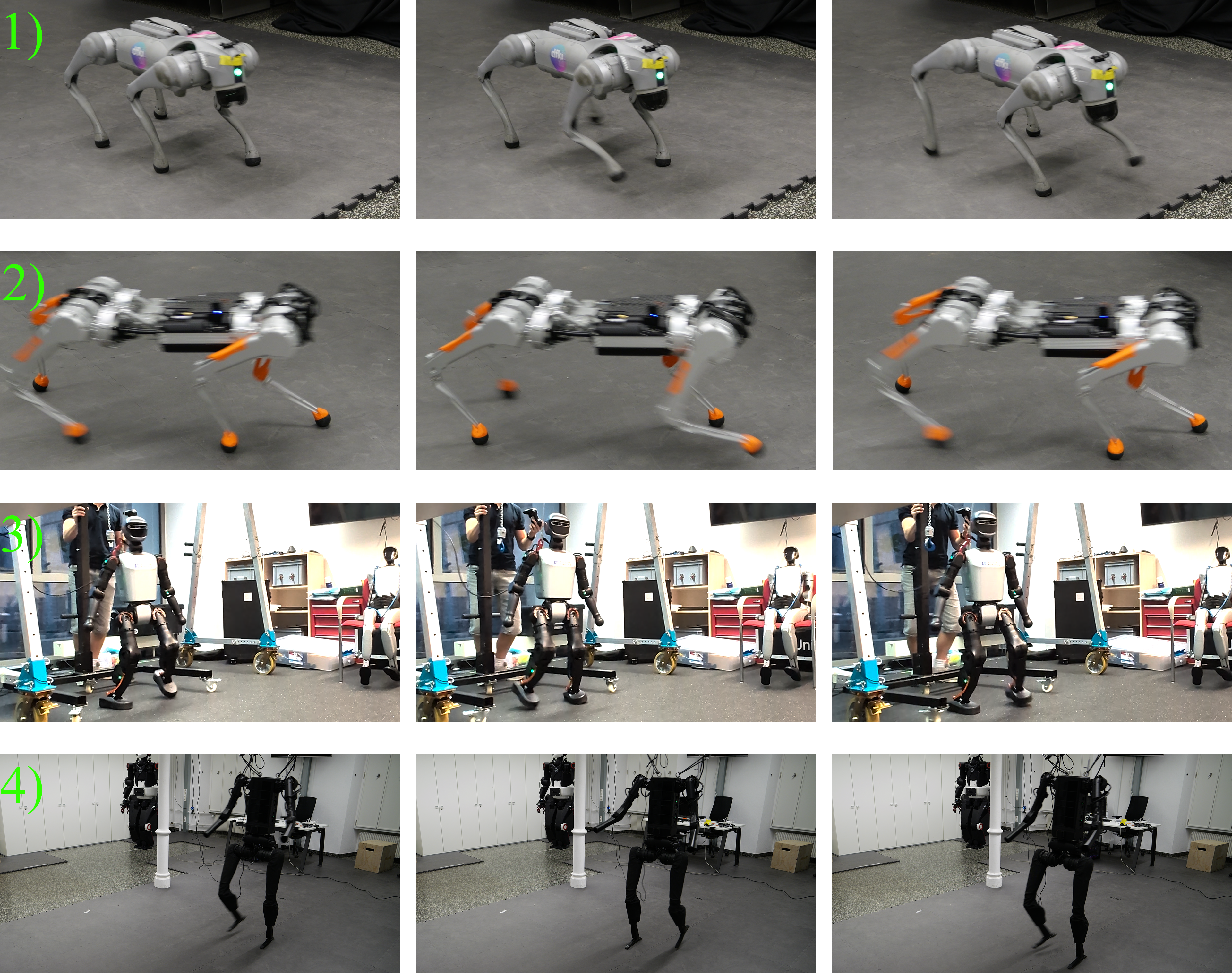}
\caption{
Shows the zero-shot transfer of \gls{urmav2} to the Unitree Go2 (row 1), MAB Silver Badger (row 2), and Booster T1 (row 3).
We also deploy the policy trained on all 50 robots on the Unitree H1 (row 4).
}
\label{fig:sim_to_real}
\end{figure}

While inspecting the learned policy in simulation can give a good indication of its performance, the ultimate test is its transfer to the real robots.
\autoref{fig:sim_to_real} shows the zero-shot sim-to-real transfer of \gls{urmav2} trained on 49 robots to the Unitree Go2, MAB Silver Badger, and Booster T1.
The policy is able to control both quadrupeds very well, even under disturbances like pushes and pulling on the legs.
The policy is able to walk forward and sidewards reliably on the Booster T1, but struggles with turning and walking backwards, leading to regular falls.
We could not zero-shot transfer to the Unitree H1 as the policy was not stable enough, but we transfered to \gls{urmav2} policy trained on all 50 robots, which was able to locomote well on the H1 in every direction.

\section{CONCLUSION}
We presented \gls{urmav2}, an improved embodiment-aware architecture for learning a general locomotion policy across a diverse set of 50 legged robots with extreme \gls{er} and a performance-based curriculum.
While \gls{urmav2} shows strong training performance and zero-shot transfer to unseen quadruped and humanoid robots in simulation, sim-to-real transfer to unseen humanoids still remains challenging.
Even more embodiment diversity through more base robots and curriculum learning methods that can explore the embodiment space more effectively could help to improve the generalization capabilities to obtain a true foundation model for robot locomotion.

\addtolength{\textheight}{-12cm}



\section*{ACKNOWLEDGMENT}

This research is funded by the National Science Centre Poland (Weave programme UMO-2021/43/I/ST6/02711), and by the German Science Foundation (DFG) (grant number PE 2315/17-1).

We gratefully acknowledge support from the hessian.AI Service Center (funded by the Federal Ministry of Education and Research, BMBF, grant no. 01IS22091) and the hessian.AI Innovation Lab (funded by the Hessian Ministry for Digital Strategy and Innovation, grant no. S-DIW04/0013/003).

We acknowledge EuroHPC Joint Undertaking for awarding us access to MareNostrum5 at BSC, Spain.


\bibliographystyle{IEEEtran}
\bibliography{IEEEabrv,bibliographie}

\end{document}